\documentclass{article}

\usepackage[preprint, nonatbib]{neurips_data_2023}
\usepackage[utf8]{inputenc} % allow utf-8 input
\usepackage[T1]{fontenc}    % use 8-bit T1 fonts
\usepackage{hyperref}       % hyperlinks
\usepackage{url}            % simple URL typesetting
\usepackage{booktabs}       % professional-quality tables
\usepackage{amsfonts}       % blackboard math symbols
\usepackage{nicefrac}       % compact symbols for 1/2, etc.
\usepackage{microtype}      % microtypography
\usepackage{xcolor}         % colors

\usepackage{graphicx} %插入图片的宏包
\usepackage{float} %设置图片浮动位置的宏包
\usepackage{CJKutf8}
\usepackage{CJK}
\usepackage{multirow}
\usepackage{tabularx}
\usepackage{caption}
\usepackage{amssymb}
\usepackage{subcaption}

\title{CC-Riddle: A Question Answering Dataset \\of Chinese Character Riddles}

\author{%
  Fan Xu \\
  Peking University \\
  Beijing, China\\
  \texttt{xufan2000@pku.edu.cn} \\
  \And
  Yunxiang Zhang \\
  University of Michigan \\
  Ann Arbor, USA \\
  \texttt{yunxiang@umich.edu} \\
  \And
  Xiaojun Wan \\
  Peking University \\
  Beijing, China\\
  \texttt{wanxiaojun@pku.edu.cn} \\
}

\begin{document}
\begin{CJK}{UTF8}{gbsn}

\maketitle

\begin{abstract}
  The Chinese character riddle is a unique form of cultural entertainment specific to the Chinese language. It typically comprises two parts: the riddle description and the solution.
  The solution to the riddle is a single character, while the riddle description primarily describes the glyph of the solution, occasionally supplemented with its explanation and pronunciation.
  Solving Chinese character riddles is a challenging task that demands understanding of character glyph, general knowledge, and a grasp of figurative language.
  In this paper, we construct a \textbf{C}hinese \textbf{C}haracter riddle dataset named CC-Riddle, which covers the majority of common simplified Chinese characters. The construction process is a combination of web crawling, language model generation and manual filtering.
  In generation stage, we input the Chinese phonetic alphabet, glyph and meaning of the solution character into the generation model, which then produces multiple riddle descriptions. The generated riddles are then manually filtered and the final CC-Riddle dataset is composed of both human-written riddles and these filtered, generated riddles. In order to assess the performance of language models on the task of solving character riddles, we use retrieval-based, generative and multiple-choice QA strategies to test three language models: BERT, ChatGPT and ChatGLM. The test results reveal that current language models still struggle to solve Chinese character riddles. CC-Riddle is publicly available at \url{https://github.com/pku0xff/CC-Riddle}.
\end{abstract}

\section{Introduction}
    \label{intro}
    % 字谜的文化背景、意义

    \begin{figure}
    \centering
    \includegraphics[width=0.6\linewidth]{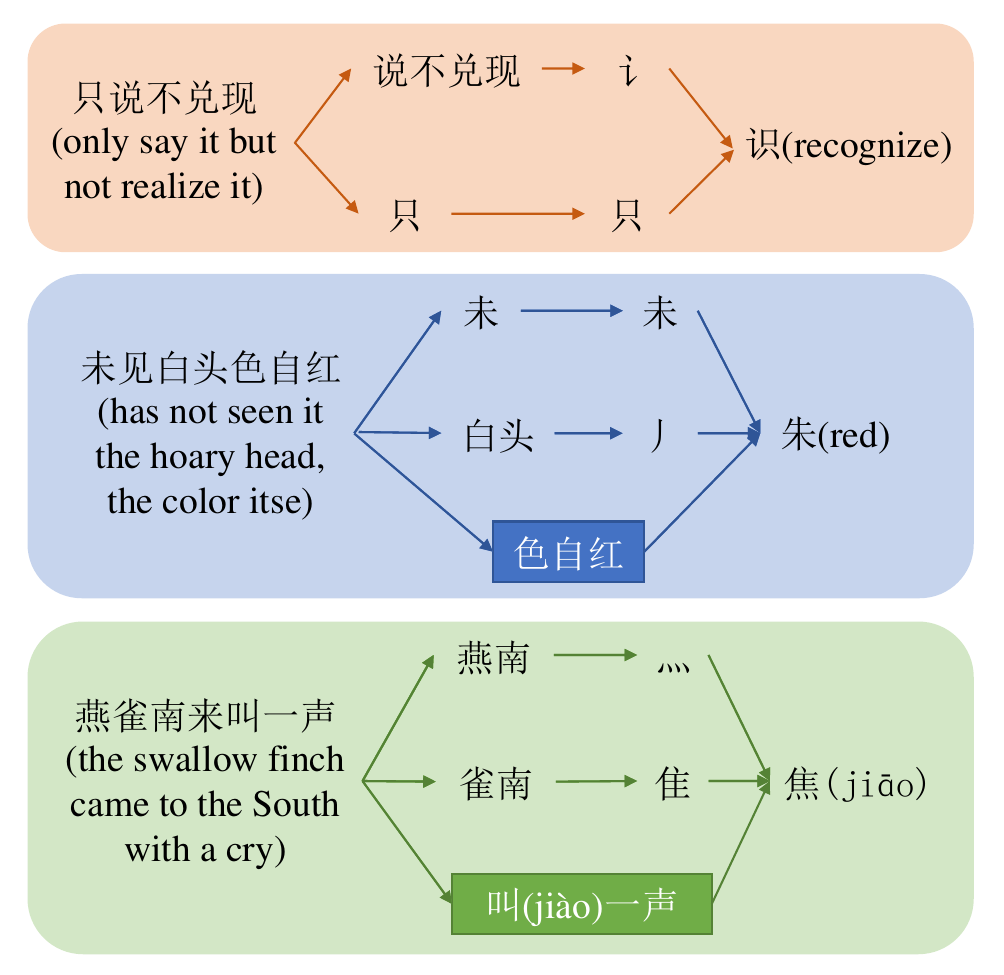}
    \caption{Three examples of character riddles.}
    \label{fig.intro}
    \end{figure}
    
    The Chinese character riddle is one of the most popular traditional riddle games in China, often appearing in traditional Chinese lantern festivals. It also takes a part in education used to help people learn about Chinese characters and culture.
    % 字谜的构成与分类
    A character riddle consists of a riddle description and a single character as its solution. The riddle description is usually created based on the glyph, explanation or pronunciation of the solution character, and uses figurative language including allusion, punning and metaphor, which makes it challenging to solve character riddles. Specifically, riddle solving also requires knowledge in the area of Chinese traditional culture.
    According to which types of information the riddle description contains, character riddles can be classified into 3 categories: glyph-based, meaning-based and pronunciation-based \cite{xiang2008chinese}. A single character riddle can have multiple categories. Figure \ref{fig.intro} demonstrates an example for each category and their interpretations are as below:
    \begin{itemize}
    \item
    \textbf{Glyph-based.}
     "识"(recognize) is composed of "讠"(say) and "只"(only), and "只说不兑现"(only say it but not realize it) is a representative description of "识". "说不兑现" means that "兑" disappears from "说", so "讠" is left over. Putting "只" and "讠" together, we obtain the answer "识".
    \item
    \textbf{Meaning-based.}
    Sometimes the description also offers hints by indicating the explanation of the character. "未见白头色自红"(has not seen the hoary head, the color itself is red) is a description of "朱"(red), where "白头"(the hoary head) indicates "丿" and "色自红"(the color itself is red) explain the meaning of "朱"(red).
    \item
    \textbf{Pronunciation-based.}
    The pronunciation of Chinese characters is represented by pinyin. The pinyin of "焦" is "jiāo". Its riddle description "燕雀南来叫一声"(the swallow finch came to the South with a cry) describes this pronunciation by changing the pinyin of "叫" from the falling tone(jiào) to the high-level tone(jiāo). 
    Meanwhile, the phrase "燕雀南来" means selecting the south part of "燕" and "雀", which presents "焦" from its glyph.
    \end{itemize}

    \iffalse
    \begin{figure}[htbp]
      \centering
      \begin{minipage}[c]{0.48\linewidth}
        \centering
        \includegraphics[width=\textwidth]{intro.pdf}
        \caption{Three examples of character riddles.}
        \label{fig.intro}
      \end{minipage}
      \hfill
      \begin{minipage}[c]{0.48\linewidth}
        \centering
        \includegraphics[width=\textwidth]{process.pdf}
        \caption{The process of constructing the CC-Riddle dataset.}
        \label{fig.process}
      \end{minipage}
    \end{figure}
    \fi

    % 数据集的构建和分析
    In this paper, we build \textbf{CC-Riddle}, a \textbf{C}hinese \textbf{C}haracter riddle dataset that covers the majority of common simplified Chinese characters. First, we crawl Chinese character riddles from the Internet. In order to cover more characters in CC-Riddle, we take a compromise way that combines model generation and manual filtering. We collect glyph, pinyin and meaning information and utilize them to finetune a pre-trained BART\cite{lewis2019bart} model as the riddle generation model. All riddle candidates generated by the fine-tuned model are gathered and filtered manually to construct the target dataset. 

    Finally, we conduct retrieval-based, generative and multiple-choice QA experiments on CC-Riddle using BERT\cite{devlin2018bert}, ChatGPT and ChatGLM\cite{du2022glm}. We also perform a manual riddle-solving test for comparison. The results indicate that language models still struggle to solve Chinese character riddles and lag far behind humans. Furthermore, our dataset can be used as a tool to evaluate how well a language model can handle figurative language, commonsense and Chinese character glyph. We hope that it will contribute to the development of more advanced Chinese and multilingual language models, and provide valuable assistance to researchers in the field of Chinese language studies.
    
    The contributions of our work are summarized as follows:
    \begin{enumerate}
        \item To the best of our knowledge, we propose and release the first large-scale, high-coverage QA dataset of Chinese character riddles, CC-Riddle.
        \item To cover more common simplified Chinese characters in our dataset, we propose a neural generation model based on BART for generating brand new riddles.
        \item We conduct a series of experiments on the CC-Riddle dataset and benchmark the results of typical neural baselines.
    \end{enumerate}

\section{Related Work}
    % 谜语数据集
    To the best of our knowledge, there has been no publicly available large-scale Chinese character riddle dataset, while datasets for common riddles have been developed before.
    % 普通谜语有什么特点？和字谜有什么关系和差别？
    Lin et al.\cite{lin2021riddlesense} present the first large dataset for answering riddle-style commonsense questions, aiming to test the abilities necessary for advanced natural language understanding(NLU). 
    Efrat et al. \cite{efrat2021cryptonite} propose a large-scale dataset based on cryptic crosswords, which is both linguistically complex and naturally sourced.
    However, these two datasets are English-only.
    Zhang and Wan \cite{zhang2021birdqa}
    introduce a bilingual dataset for QA on riddles that contain both Chinese and English riddles.
    For the Chinese part, they exclude Chinese character riddles and focus on common Chinese riddles which imply the meaning of the answer.
    
    % 字谜/谜语生成
    Chinese character riddle generation has been explored before with statistical method. Tan et al. \cite{tan2016solving} propose a method to solve and generate Chinese character riddles by extracting "radical-phrase" alignments and rules from existing character riddles. Template-based method and replacement-based method are the two ways to generate riddles. Then Ranking SVM is used to rerank the generated candidates. Unfortunately, the data and code of this work cannot be accessed publicly. Preliminary studies have been taken on the common riddle as well. 
    Binsted and Ritchie \cite{binsted1994implemented,binsted1994symbolic} implement a program to generate riddles from humour-independent lexical entries, named JAPE-1. Galvan et al. \cite{galvan2016riddle} utilize the common properties among different concepts and existing knowledge bases to generate riddles by comparisons.
    
    % 其他相关工作：关于古诗、对联、汉字
    Neural network has not been used to generate Chinese character riddles, but has been used to generate Chinese poems and couplets widely \cite{yi2017generating,chiang2021transcouplet}. There are researches focusing on Chinese character as well. Some are around Chinese character embedding models   \cite{li2015component,lu2016multi,tseng2019eigencharacter}. The glyph information of Chinese characters can also be incorporated into language model pre-training \cite{li2021glyphcrm,sun2021chinesebert}.

\section{Dataset Construction and Analysis}
    \begin{figure}
    \centering
    \includegraphics[width=\linewidth]{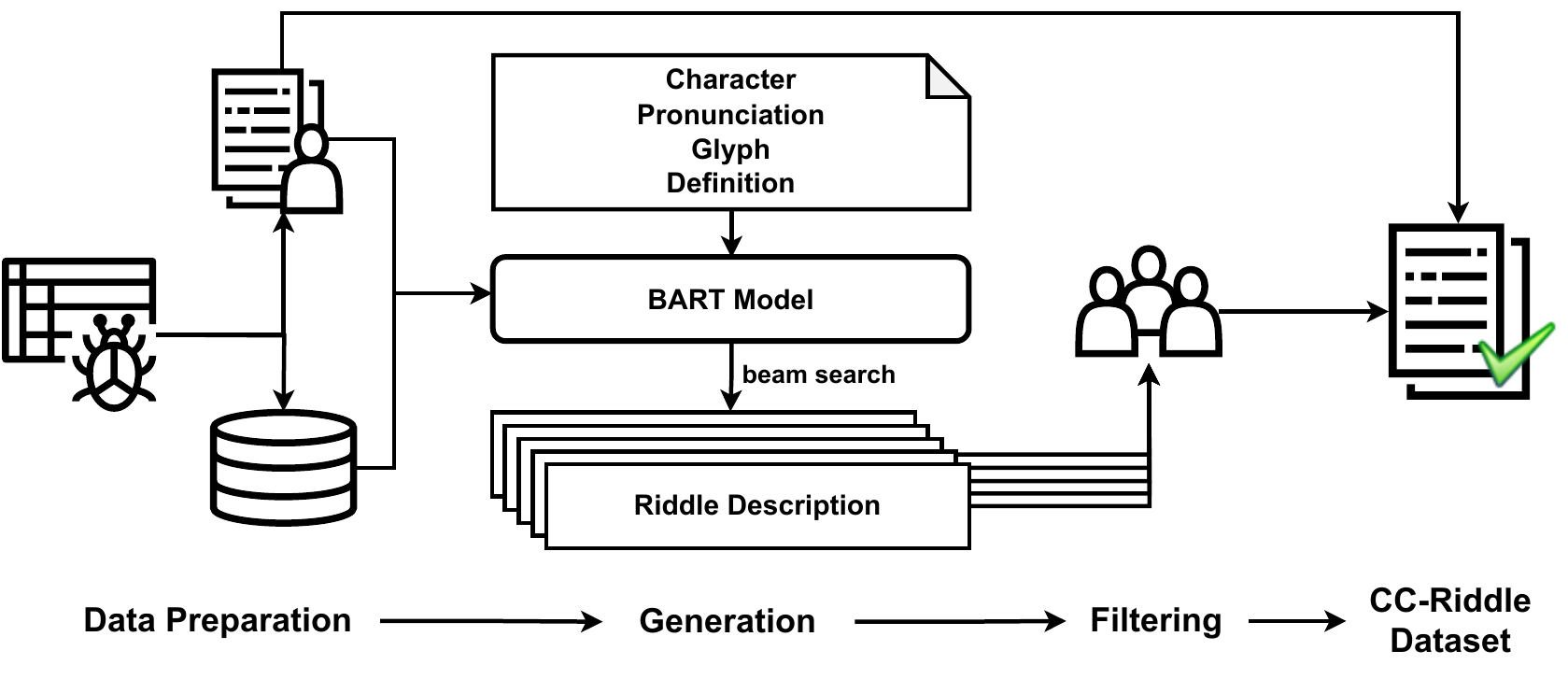}
    \caption{The process of constructing the CC-Riddle dataset.}
    \label{fig.process}
    \end{figure}
    In this section, we introduce our method to construct the CC-Riddle dataset in detail (Figure~\ref{fig.process}). The dataset includes Chinese character riddles written by humans and generated by machines. First, human-written character riddles are crawled from the Web. However, some frequently used Chinese characters are not included in the crawled data, so new character riddles are needed to build a large-scale dataset with a high coverage. Manual design and automatic generation are two ways to obtain more Chinese character riddles other than crawling from the Web. However, creating riddles manually needs a great deal of time and requires high literacy of the writers. Meanwhile, existing generative models are still weak in understanding Chinese character riddles. Therefore, we take a compromise approach that combines generating automatically and filtering manually to acquire reasonable Chinese character riddles at a relatively low cost.

\subsection{Data Preparation}
    \label{data_preparation}
    We scrape existing Chinese character riddles from some websites\footnote{\url{http://xh.5156edu.com/}, \url{https://www.chnlib.com/miyuku/zimi/index.html}, \url{https://www.cmy123.com/zmmy}, \url{http://www.hydcd.com/baike/zimi.htm}, \url{https://www.caimigu.com/zm/}}, remove duplicate data and delete irrelevant characters to enhance data quality.
    After this data cleaning process, we acquire a total of 20,385 unique Chinese character riddles. These riddles encompass 5,060 Chinese characters in total, the majority of which are frequently used.
    However, some common Chinese characters are still not included in the crawled data. More Chinese character riddles are needed to improve the coverage of CC-Riddle. The pronunciation, glyph and explanation of the solution character are important for generating brand new Chinese character riddles \cite{xiang2008chinese}. Therefore, we obtain glyph information from the IDS dictionary\footnote{\url{https://github.com/cjkvi/cjkvi-ids}} and HanziCraft\footnote{\url{https://hanzicraft.com/}}, as well as pinyin and meaning from the Chinese Xinhua Dictionary\footnote{\url{https://github.com/pwxcoo/chinese-xinhua}}.
    The IDS dictionary is a resource about the glyph of characters. It provides a coarse-grained decomposition of characters and describes the position relationship between different components of a character (i.e., the character structure).
    HanziCraft is a web application that provides information about Chinese characters, in which there are three kinds of character decomposition: once, radical and graphical. Figure \ref{fig.decomposition} shows some examples from IDS dictionary and HanziCraft.
    % We use decomposition with structure from IDS and radical decomposition from HanziCraft as the glyph information.

    \iffalse
    \begin{figure}
    \centering
    \includegraphics[width=\linewidth]{structure.pdf}
    \caption{Ten types of Chinese character structures.}
    %\Description{Single-radical, left-right, up-down, up-left, up-right, left-down, up-left-down, left-up-right, left-down-right and enclosure.}
    \label{Fig.main2}
    \end{figure}
    \fi

    % CJK包无法表示一些特殊字符，所以把表做成图了

\subsection{The Generation Model}
    \label{generation}

    \iffalse
    \begin{figure}
    \centering
    \includegraphics[width=0.6\linewidth]{gen_model.pdf}
    \caption{Illustration of the generation model during inference.}
    \label{fig.gen_model}
    \end{figure}
    \fi
    
    Upon collecting the requisite data as previously specified, we design and subsequently train a generative model to produce brand new Chinese character riddles. Particularly, we fine-tune a pre-trained Chinese BART model\footnote{\url{https://huggingface.co/fnlp/bart-base-chinese}} \cite{shao2021cpt} using the Transformers library\cite{wolf2020transformers}. 
    During training, our model takes a Chinese character together with all the three elements (pinyin, glyph and explanation) as input. We use decomposition from IDS dictionary and radical decomposition from HanziCraft as glyph information. The output is a single riddle description. One character may have multiple riddle descriptions and each of them acts as an independent piece of data in training.
    Training set, validation set and test set are split by characters.
    %so that we can avoid generating riddles highly similar to existing riddles and evaluate the generalization ability of the model effectively. 
    We use 98 characters for the validation set and 100 characters for the test set. The rest are all included in the training set. 
    As for hyperparameters, we set learning rate to \texttt{5e-5}, batch size \texttt{16}, random seed \texttt{42} and number of epochs \texttt{12}.
    In the generation stage, the solution character is prevented to appear in generated riddle descriptions by setting its \texttt{logit} to \texttt{-inf} in decoding. 
    We use beam search as the decoding algorithm to generate 5 riddle description candidates for each input entry. The generation part of Figure \ref{fig.process} illustrates the process of generating Chinese character riddles.
    Besides, we implement the template-based method and replacement-based method proposed in \cite{tan2016solving} with our data and take them as the baselines.\footnote{While implementing the baselines, we could not train the weights of features for the ranking model because the data for training it is annotated by human and not publicly available. We make a rough adjustment on the weights manually instead.}
    
    \begin{table}
    \caption{The BLEU-4 scores and numbers of distinct unigram (\# dist. unigrams) and bigrams (\# dist. bigrams) in the top-1 riddle descriptions in test set.}
    \label{tab:gen_eval}
    \centering
    \setlength\tabcolsep{8pt}
      \begin{tabular}{l|c|cc}
        \toprule
        \multirow{2}*{\textbf{Method}} & \multirow{2}*{\textbf{BLEU-4}} & \textbf{\# Distinct} & \textbf{\# Distinct}   \\
        ~ & & \textbf{unigrams} & \textbf{bigrams} \\
        \midrule
        Ours & 3.87 & \textbf{353} & \textbf{612} \\
        \midrule
        Template-based & 4.67 & 128 & 280 \\
        Replacement-based & \textbf{29.33} & 248 & 343 \\
      \bottomrule
    \end{tabular}
    \end{table}

    We adopt several automatic metrics to evaluate the generation results, including BLEU-4\cite{papineni2002bleu} and the number of distinct n-grams. The evaluation results are shown in Table \ref{tab:gen_eval}.
    We first report the BLEU scores on 100 tested characters for the top-1 outputs of our model and the baselines. 
    The BLEU scores of our model and template-based baseline are very low because many different reasonable riddle descriptions may be generated for a Chinese character but they do not need to be lexically similar to the given reference descriptions.
    The replacement-based method takes human-written riddles as direct reference and only a few phrases are replaced in the generated ones, so the BLEU score is high. Therefore, the BLEU scores are only for reference and it is not meaningful to compare the models based on BLEU.
    To assess the diversity of the generated descriptions, we calculate the numbers of distinct uni-grams and bi-grams. The results suggest that our method far exceeds baselines in diversity, which is important for such riddle games.

\subsection{Filtering}
    \label{filtering}
    % TODO: 保证/评估数据质量？
    % TODO: 一致性衡量，Pearson correlation or Cohen's kappa

    \begin{figure}[htbp]
    \centering
    \begin{minipage}[c]{0.50\textwidth}
        \begin{table}[H]
            \caption{The criteria for the classification of acceptable candidate riddles. When the number of label grows, the riddle description becomes more related to the meaning and less related to the glyph of the answer character.}
            \label{tab:classify}
            \centering
            \setlength\tabcolsep{3pt}
            \begin{tabular}{cl}
            \toprule
            \textbf{Label} & \textbf{Criterion} \\
            \midrule
            1 & The riddle description can describe the \\
              & shapes or meanings of all or most com- \\
              & ponents for the answer character. \\
              & e.g. 西湖波底映花前——菠 \\
            \midrule
            2 & The riddle description can describe the \\
              & shapes or meanings of part components \\
              & of the answer character, and at the \\
              & same time express the meaning of the \\
              & character. \\
              & e.g. 匡王之言不可信——诓  \\
            \midrule
            3 & The riddle description can describe the \\
              & meaning or use case of the answer \\
              & character. \\
              & e.g. 戏言——谑 \\
            \bottomrule
            \end{tabular}
        \end{table}
    \end{minipage}
    \hfill
    \begin{minipage}[c]{0.45\textwidth}
        \includegraphics[width=\textwidth]{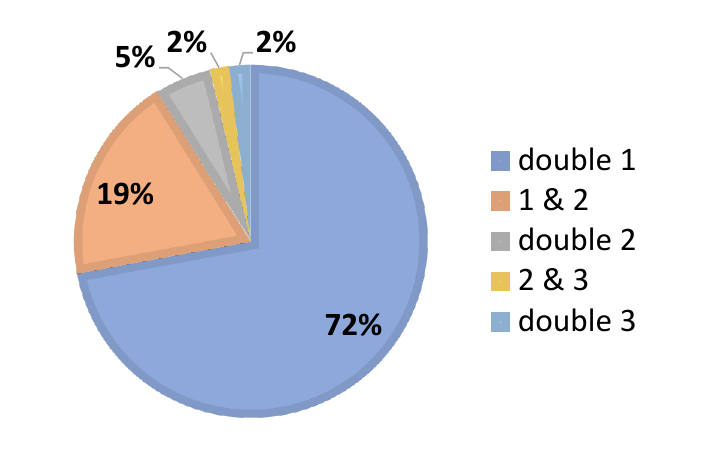}
        \caption{The proportions of each category of accepted riddle candidates. Each accepted riddle description is categorized by the two annotators who deem it to be reasonable. "Double $X$" means that both annotators label it as $X$. "$X \& Y$" means that one annotator labels it as $X$ and the other one labels it as $Y$. The labels in this figure correspond to those in Table \ref{tab:classify}.}
        \label{fig.classify}
    \end{minipage}
    \end{figure}

    After generation, we get 59327 different candidates in total for 4293 characters out of the training set. Then they are manually filtered to construct CC-Riddle. For each candidate, we assign it to two annotators at first. The agreement ratio is 83.95\% on whether to accept the candidates.
    % agree to accept: 5129 / 59327 = 0.08645304835909451
    % agree to reject: 44674 / 59327 = 0.7530129620577477
    % agree: 49803 / 59327 = 0.8394660104168422
    % disagree: 9524 / 59327 = 0.16053398958315776
    If the first two annotators fail to agree on the acceptance of this candidate, then the third one will make the decision. 
    If a candidate can be categorized into one of the three categories shown in Table \ref{tab:classify}, it is acceptable. Here we don't take the pronunciation into consideration because pinyin-based riddles are rare in human-written riddles and are difficult to generate without special processing.
    The annotator needs to decide whether the candidate is acceptable and how the riddle description reveals the answer according to the information it contains at the same time. We hire 20 annotators in total and all of them are native Chinese speakers and are undergraduates or have better qualifications. 
    % Each annotator is compensated with 0.1 yuan for every candidate they review.

    \iffalse
    \begin{table}
    \caption{Examples of generated character riddles with different labels.}
    \label{tab:gen_example}
    \centering
    \begin{tabular}{lll}
    \toprule
    Label & Character & Riddle description\\
    \midrule
    \multirow{3}*{1} & 违 & 伟人走了送出关\\
    ~ & disobey & the great man left and was \\
    ~ & ~ & sent out at the gateway \\
    \midrule
    \multirow{2}*{2} & 擤 & 用手揉捏鼻子\\
    ~ & blow & knead the nose with hands \\
    \midrule
    \multirow{2}*{3} & 假 & 不是真的\\
    ~ & fake & be not true \\
    \bottomrule
    \end{tabular}
    \end{table}
    \fi
    
    Finally, we get 7132 generated riddles after manual filtering. The proportions of each category are shown in Figure \ref{fig.classify}. Obviously, most character riddles are based on the glyph of its solution. Only a small part of them make use of the meaning and the riddles that uses meaning purely are even less. This is because most riddles in the training set are glyph-based ones.
    
\subsection{Dataset Analysis}

    \begin{table}[ht]
    \begin{minipage}[c]{0.48\textwidth}
    \centering
    \setlength\tabcolsep{2pt}
    \caption{Coverage of CC-Riddle on two formal Chinese character sets.}
    \label{tab:coverage}
    \begin{tabular}{l|cccc}
    \toprule
    \multirow{2}*{\textbf{Character det}} & \multirow{2}*{\textbf{Size}} & \multirow{2}*{\textbf{Cov.}} & \textbf{Cov.} & \textbf{Cov.}\\
    &&& \textbf{by gen.} & \textbf{rate} \\
    \midrule
    GB/T 2312-1980  & 6763 & 6075 & 1312 & 0.8983 \\
    TGSCC           & 8105 & 6194 & 1459 & 0.7642 \\
    TGSCC-frequent  & 3500 & 3358 & 332  & 0.9594 \\
    \bottomrule
    \end{tabular}
    \end{minipage}
    \hfill
    \begin{minipage}[c]{0.50\textwidth}
    \centering
    \setlength\tabcolsep{2pt}
    \caption{Basic statistics of CC-Riddle.}
    \label{tab:statistics}
    \begin{tabular}{l|ccc}
    \toprule
    \textbf{Split} & \textbf{\# Characters} & \textbf{\# Riddles} & \# \textbf{Riddles(gen.)} \\
    \midrule
    train & 4367 & 16626 & 4165 \\
    valid & 1457 & 5480 & 1387 \\
    test & 1455 & 5411 & 1580 \\
    \textbf{total} & \textbf{7279} & \textbf{27517} & \textbf{7132} \\
    \bottomrule
    \end{tabular}
    \end{minipage}
    \end{table}
    
    \paragraph{Size and Coverage.}
    The CC-Riddle dataset integrates generated Chinese character riddles with web-crawled ones, culminating in a comprehensive collection of 27,517 unique riddles (Table \ref{tab:statistics}). It covers 7,279 characters and most frequently used characters are included. The whole dataset is split into training set, validation set and test set by characters 
    %(i.e., a character will not appear in more then one set at the same time as an answer.) 
    with a ratio of 6:2:2. The ratios of generated riddles to web-crawled ones are roughly the same across training, validation and test set.
    We compare our dataset with two formal sets of Chinese characters: GB/T 2312-1980\footnote{A key official character set of China, used for simplified Chinese characters.} and the Table of General Standard Chinese Characters(TGSCC)\footnote{The current standard list of 8,105 Chinese characters published by the government of the People's Republic of China.}. Among all the 8,105 characters in TGSCC, 3,500 are designated as frequent. CC-Riddle covers most characters in these sets, as is shown in Table \ref{tab:coverage}.
    There are two main reasons that some characters are missing from the CC-Riddle. First, generated riddles for these characters are of low quality and are all rejected during filtering. In most cases, descriptions of key radicals are absent in the rejected riddle. Second, some characters in GB/T 2312-1980 are radicals themselves and necessary information (i.e., pinyin, decomposition and explanation) of them is missing.
    
\begin{figure}[htbp]
      \centering
      \begin{minipage}[c]{0.48\linewidth}
        \centering
        \includegraphics[width=\textwidth]{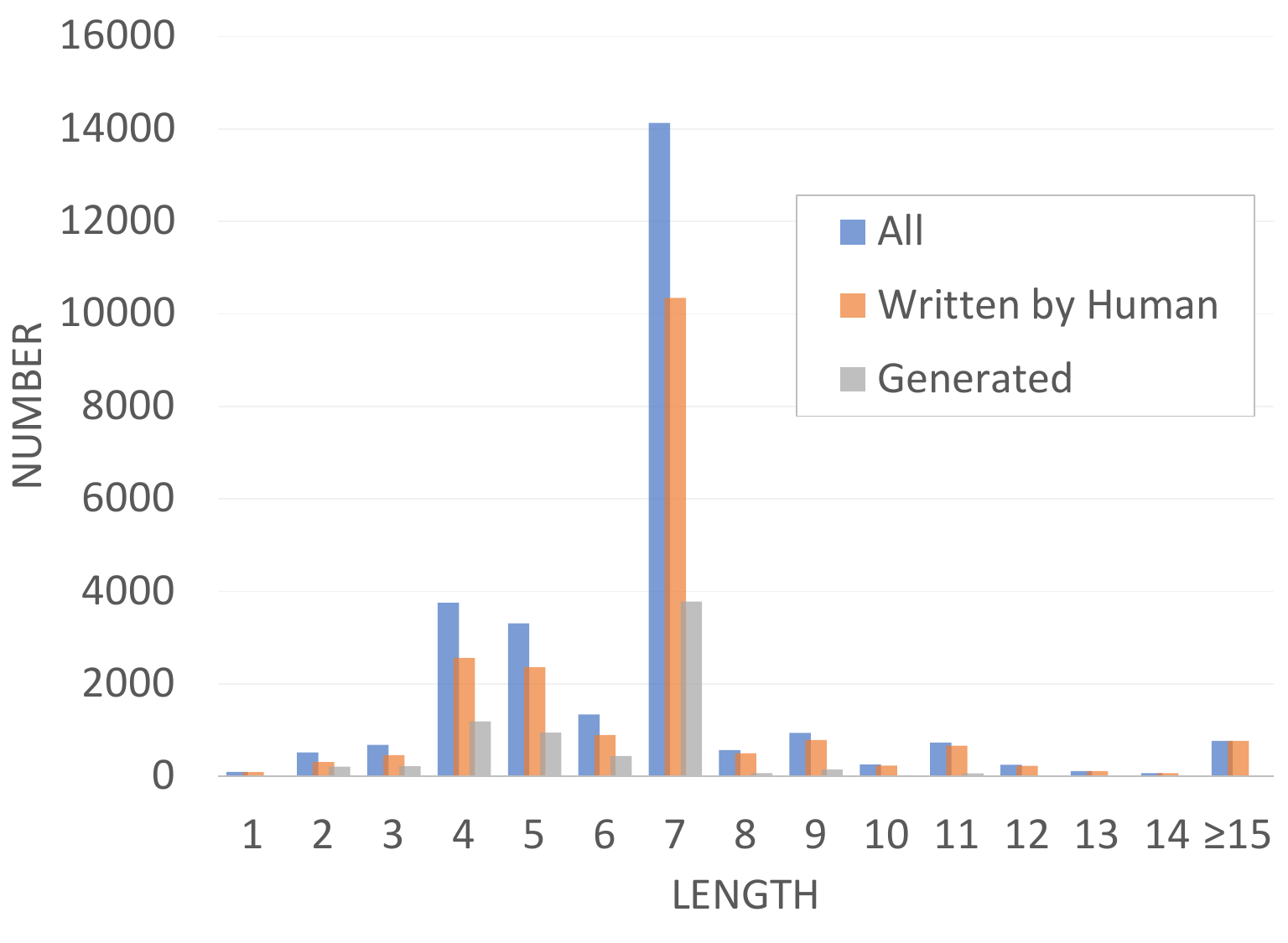}
        \caption{The distribution of riddle description length.}
        \label{fig.length}
      \end{minipage}
      \hfill
      \begin{minipage}[c]{0.48\linewidth}
        \centering
        \includegraphics[width=\textwidth]{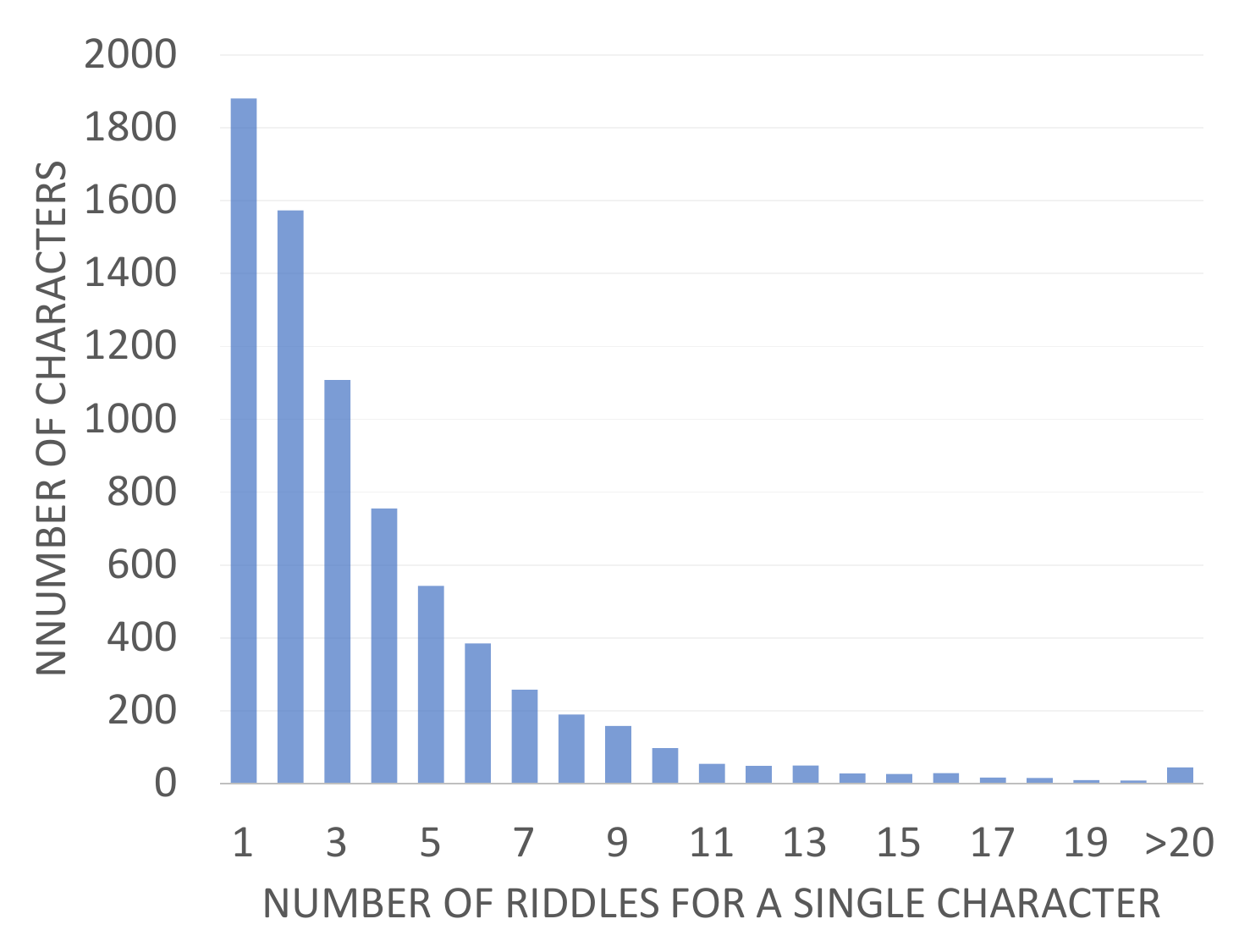}
        \caption{The distribution of multiplicity: Every solution character can be associated with multiple riddle descriptions.}
        \label{fig.multiplicity}
      \end{minipage}
    \end{figure}

    \paragraph{Length.}
    Unlike Chinese poems or couplets, Chinese character riddles aren't bound by any rules pertaining to the length of their descriptions. However, an interesting observation is that the majority of these riddles tend to be rather short. Figure \ref{fig.length} shows the distribution of the character riddle length. The generated riddles and human-written ones are similar in distribution while almost all long riddles are human-written ones.
    Most riddles are at a length of 7 characters, which is also the typical length of a sentence in traditional Chinese poetry.
    Riddles whose length exceed 16 often come in the form of a poem.
    For example, a long riddle description for "思" (miss) is "清波映月绕山峦，南亩扶犁白露沾；息隐冈中须蓄志，一心要把四方安" (clear waves reflect the moon around the mountains, and the plow in the south field is stained with white dew; resting in the hidden hill, he is prepared and determined to put the country in peace). 

    \paragraph{Multiplicity.}
    Many solution characters in CC-Riddle have multiple corresponding riddle descriptions. Figure \ref{fig.multiplicity} shows the distribution of the number of riddle descriptions. Most characters have only one riddle description, and generally, the number of solution characters dramatically declines as the number of riddle descriptions increases. Very few characters have more than 10 riddle descriptions. Typically, more commonly used characters have more riddle descriptions. The character "一" has the most riddle descriptions, with a total of 177.

    \paragraph{Frequent Words.}
    Some words frequently appear in riddle descriptions and they can partly represent the characteristics of Chinese character riddles. We split the riddle descriptions into words using the Chinese word segmentation module Jieba\footnote{\url{https://github.com/fxsjy/jieba}} and count up the number of occurrences of the words. Table \ref{tab:freq} shows the top-15 most frequent words. Some of them indicate common radicals such as "西湖"(氵). Some of them indicate operations such as "一半"(a half). It can also be inferred from these words that punning is widely used in Chinese character riddles. For instance, "四方" could be interpreted as four squares, while "分" could denote a unit of length or time. In addition, imageries such as "西湖" and "月" (moon) frequently appear in traditional Chinese poetry, suggesting a stylistic resemblance between some Chinese character riddles and traditional Chinese poetry.

    \begin{table}[H]
    \caption{The top-15 most frequent words in riddle descriptions.}
    \label{tab:freq}
    \centering
    \setlength\tabcolsep{6pt}
    \begin{tabular}{ll|ll|ll|ll|ll}
    \toprule
    \textbf{Word} & \textbf{Freq.} & \textbf{Word} & \textbf{Freq.} & \textbf{Word} & \textbf{Freq.} & \textbf{Word} & \textbf{Freq.} & \textbf{Word} & \textbf{Freq.}\\
    \midrule
    一点 & 300 &
    改革 & 281 &
    一半 & 257 &
    之后 & 248 &
    花前 & 224 \\
    \multicolumn{2}{l|}{one point} &
    \multicolumn{2}{l|}{reform} & 
    \multicolumn{2}{l|}{a half} & 
    \multicolumn{2}{l|}{after that} & 
    \multicolumn{2}{l}{before flowers} \\
    \midrule
    四方 & 188 &
    个个 & 185 &
    残月 & 167 &
    不见 & 165 &
    千里 & 151 \\
    \multicolumn{2}{l|}{four directions} &
    \multicolumn{2}{l|}{every} &
    \multicolumn{2}{l|}{waning moon} &
    \multicolumn{2}{l|}{disappear} &
    \multicolumn{2}{l}{a thousand miles} \\
    \midrule
    十分 & 149 &
    一片 & 148 &
    西湖 & 145 &
    一直 & 144 &
    新月 & 137 \\
    \multicolumn{2}{l|}{very}    &  
    \multicolumn{2}{l|}{a piece} &
    \multicolumn{2}{l|}{the West Lake}   &
    \multicolumn{2}{l|}{straight}    &
    \multicolumn{2}{l}{crescent}   \\
    \bottomrule
    \end{tabular}
    \end{table}

\section{Benchmark Experiments}

    \subsection{Settings}
    \label{test_settings}

    \begin{figure}[htbp]
    \centering
    \begin{minipage}[c]{0.53\textwidth}
        \begin{table}[H]
        \centering
        \caption{The input formats of three testing methods, taking “孔子登山——岳' (Confucius ascends the mountain —— Mount) as an example. We use \texttt{[structure]} to denote the ideographic description characters.}
        \label{tab:test_sample}
        \setlength\tabcolsep{3pt}
        \begin{tabular}{lll}
        \toprule
        \multicolumn{2}{l}{\textbf{Testing method}} & \textbf{Input text} \\
        \midrule
        \multirow{6}*{Retrival} & \multirow{2}*{Character} & 孔子登山 \\
        && 岳 \\
        \cmidrule{2-3}
        & \multirow{2}*{Glyph} & 孔子登山 \\
        && 岳，\texttt{[up-down]}丘山 \\
        \cmidrule{2-3}
        & \multirow{2}*{Definition} & 孔子登山 \\
        && 岳，（会意。从山，从丘... \\
        \midrule
        \multicolumn{2}{l}{Generative QA} & 孔子登山（打一字） \\
        \midrule
        && 孔子登山（打一字）\\
        \multicolumn{2}{l}{Multiple-choice QA} & 选项：岳，晾，僦，啭  \\
        && 请选出正确选项。 \\
        \bottomrule
        \end{tabular}
        \end{table}
    \end{minipage}
    \hfill
    \begin{minipage}[c]{0.45\textwidth}
        \includegraphics[width=\textwidth]{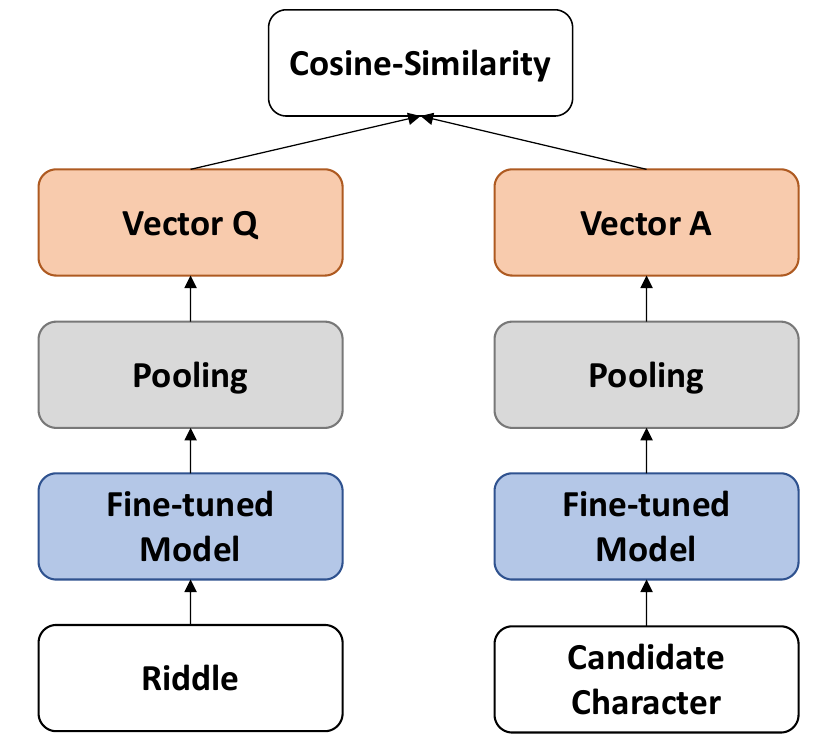}
        \caption{The process of calculating the similarity between the query (riddle description) and candidates (solution).}
        \label{fig.retrieval}
    \end{minipage}
    \end{figure}
    
    We design three QA strategies - retrieval-based, generative and multiple-choice QA - to test language models, including BERT\cite{devlin2018bert}, ChatGPT\footnote{\url{https://chat.openai.com/}} and ChatGLM\cite{du2022glm}. Table \ref{tab:test_sample} displays the same test sample under different testing methods. We also conduct manual test on QA task.
    
    \paragraph{Retrieval-based QA.}
    \label{retrieval}
    The retrieval task is designed to accommodate models built on the BERT architecture. Its format is as follows: Given the riddle description and a candidate character set, supplemented with glyph or definition, the candidate characters are sorted according to their similarity with the description.
    There is also a setting of characters without supplementary.
    We finetune a pretrained Chinese BERT model\footnote{\url{https://huggingface.co/bert-base-chinese}}\cite{devlin2018bert} in these experiments.
    All solution characters contained in the CC-Riddle dataset forms the candidate character set. Supplementary information comes from the IDS dictionary or character definition from the Xinhua Dictionary.
    % The reason we used the IDS dictionary instead of other glyph decomposition methods is that the IDS dictionary provides the relative position relationship between decomposed components in the form of special characters, which is not given in other Chinese character decomposition methods.
    The architecture used in the experiment is shown in Figure \ref{fig.retrieval}.
    The model first embeds the input text, then applies mean pooling to obtain vectors Q and A and finally calculates the cosine similarity between the two vectors.
    As is shown in Table \ref{tab:test_sample}, each input text has 2 lines, corresponding to the two input ends of the model architecture. During the training phase, we construct positive and negative samples in a 1:2 ratio.
    %For each positive training data in the form of "riddle description - solution", we randomly sample a riddle description that does not match the original solution and a character that does not match the original description to construct the two negative samples. 
    The loss function is the multiple negatives ranking loss, which is used to optimize the model's ranking ability. We train the model for 3 epochs with a batch size of 32 using the SentenceTransformers\cite{reimers2019sentence} library.
    % Compared with the Cross-Encoder structure, the Bi-Encoder structure adopted in our work can improve the retrieval efficiency during the testing phase. At the beginning of the testing phase, the model first maps all the candidates into a set of vectors in the target space. When a query (i.e., the riddle description) is inputted, the model only needs to map the riddle clue into the vector space.
    % We utilize the Mean Reciprocal Rank (MRR) as the evaluation metric and calculate the metric only for the top 1 and 5 results returned, denoted as MRR@1 and MRR@5 respectively.
    We calculate accuracy as the evaluation for the top 1 results returned.

    \paragraph{Generative QA.}
    The format of the generative QA test is as follows: input the riddle description as question and output the solution as answer. This test is consistent with the traditional riddle game format and is the most challenging among the three testing methods. We tested two popular conversational language models that perform well in Chinese, ChatGPT(\texttt{gpt-3.5-turbo-0301}) and ChatGLM-6B. Table \ref{tab:test_sample} demonstrates the prompts. We use regular expressions to process the model's output and locate the answer. This test uses accuracy as the evaluation metric.

    \paragraph{Multiple-choice QA.}
    For the multiple-choice QA test, we feed the riddle description and four candidate options to the models (Table \ref{tab:test_sample}), ChatGPT and ChatGLM, which then select the correct option. Due to the limited choice range, multiple-choice QA is the simplest of the three testing methods and can be easily solved by humans. Of the four candidate options, one is the correct solution, while the other three serve as distractors. We design these distractors in two ways: The first is to randomly sample three characters different from the solution from the entire dataset; the second is to map the riddle description to an embedding vector, select the three most similar riddles with different solutions, and use their solutions as distractors. These two design methods are intended to offer varying degrees of challenge. We employ the \texttt{text-embedding-ada-002} embedding model published by OpenAI, which can map the input text to a 1536-dimensional vector. Cosine similarity is the distance measure, and accuracy is the evaluation metric.

    \paragraph{Manual riddle-solving annotation.}
    \label{human_test}
    We also conduct a manual test to evaluate human performance on this task. We randomly sample 100 entries from the test set and hire three annotators to solve them. Two of these annotators, referred to as A and B, are senior undergraduates majoring in Chinese linguistics. The third, annotator C, is a postgraduate student majoring in history. We set the task as a closed-book exam, with the objective of solving as many riddles as possible within two hours. If any of the riddles have not been considered within the period, we extend the time until all remaining riddles have been reviewed at least once. Annotator A provides a more detailed analysis, identifying which riddles are comprehensible given the solution.
    
    \subsection{Results and Analysis}
    
    The experiment results are presented in Table \ref{tab:results}. Overall, solving Chinese character riddles is a challenging task for both language models and humans, yet the performance of language models significantly lags behind that of human annotators. When testing the models, the difficulty level progresses from multiple-choice QA, to retrieval-based QA, and finally to generative QA. The difficulty of the generative QA test is equivalent to that of the manual test. It can be inferred from the comparison between retrieval-based and generative QA tests that the supplementary definitions and glyphs contribute to solving riddles, with glyphs being especially useful.

    \begin{table}[H]
    \centering
    \caption{Overall results. We provide metrics on all test riddles, web-crawled ones and generated ones in order, separated by '/'.}
    \label{tab:results}
    
    \begin{subtable}{0.48\textwidth}
        \centering
        \caption{Retrieval-based QA test using BERT model.}
        \label{tab:retrieval}
        \setlength\tabcolsep{5pt}
        \begin{tabular}{cccc}
            \iffalse
            {
            \toprule
            \textbf{Supplement} & \textbf{MRR@5} \\
            \midrule
            Glyph & 0.1440\ /\ 0.0507\ /\ 0.3701 \\
            Definition & 0.0547\ /\ 0.0302\ /\ 0.1141 \\
            \bottomrule
            }
            \fi
            \toprule
            \multirow{2}*{\textbf{Supplement}} & \multicolumn{3}{c}{\textbf{Accuracy}} \\
            ~ & all & web & gen  \\
            \midrule
            Character & 0.0015 & 0.0016 & 0.0013 \\
            Glyph & 0.1133 & 0.0321 & 0.3101 \\
            Definition & 0.0381 & 0.0170 & 0.0892 \\
            \bottomrule
        \end{tabular}
    \end{subtable}
    ~
    \begin{subtable}{0.48\textwidth}
        \centering
        \caption{Generative QA test.}
        \label{tab:qa}
        \setlength\tabcolsep{5pt}
        \begin{tabular}{cccc}
            \toprule
            \multirow{2}*{\textbf{Model}} & \multicolumn{3}{c}{\textbf{Accuracy}} \\
            ~ & all & web & gen  \\
            \midrule
            ChatGPT & 0.0039 & 0.0044 & 0.0025 \\
            ChatGLM & 0.0013 & 0.0016 & 0.0006 \\
            \bottomrule
        \end{tabular}
    \end{subtable}
    \\
    \begin{subtable}{0.48\textwidth}
        \centering
        \caption{Multiple-choice QA test.}
        \setlength\tabcolsep{3pt}
        \label{tab:multiplechoice}
        \begin{tabular}{ccccc}
            \toprule
            \multirow{2}*{\textbf{Option}} & \multirow{2}*{\textbf{Model}} & \multicolumn{3}{c}{\textbf{Accuracy}} \\
            ~ & ~ & all & web & gen  \\
            \midrule
            \multirow{2}*{Random} & ChatGPT & 0.4153 & 0.3816 & 0.4968 \\
            & ChatGLM & 0.2706 & 0.2999 & 0.1994 \\
            \multirow{2}*{Similar} & ChatGPT & 0.4016 & 0.3587 & 0.5057 \\
            & ChatGLM & 0.2277 & 0.2597 & 0.1500 \\
            \bottomrule
        \end{tabular}
    \end{subtable}
    ~
    \begin{subtable}{0.48\textwidth}
        \centering
        \caption{Manual annotation on 100 test samples.\protect\footnotemark}
        \label{tab:human}
        \begin{tabular}{cccc}
            \toprule
            \multirow{2}*{\textbf{Annotator}} & \multicolumn{3}{c}{\textbf{Accuracy}} \\
            ~ & all & web & gen  \\
            \midrule
            A & 0.38 & 0.43 & 0.27 \\
            B & 0.18 & 0.22 & 0.09 \\
            C & 0.65 & 0.87 & 0.21 \\
            \bottomrule
        \end{tabular}
    \end{subtable}
    \end{table}

    \footnotetext{Annotator C has discussed some of the tested riddles with their workmates.}
    
    \paragraph{Errors and difficulties.}
    We observe that there are three main error types: incomprehension, misunderstanding and alternative solutions.
    Out of the 100 manually annotated riddles, 38 are correctly solved and 3 are annotated with multiple solutions. Even after we provide all solutions to the annotator, 31 riddles remain incomprehensible, 12 of which are generated riddles. For models, most of the failures can be attributed to incomprehension and misunderstanding. The most challenging part in riddle solving is matching the phrases in the description to correct radicals. Upon examining the test results, we find that there are 1470 riddles that none of the models can handle correctly, comprising 1078 web-crawled and 392 generated riddles.

    \label{analysis}
    \paragraph{Comparative analysis of web-crawled and model-generated riddles.}
    For models, the level of difficulty varies between the two, depending on the specific model and the testing method. For humans, the riddles generated by the model are more challenging to solve. We've held brief discussions with the annotators and summarize the relationships and differences between the two types: First, some generated riddles are modified from existing ones; second, manually designed riddles are often more readable, and can even form poetic phrases or contain philosophical implications, which are still difficult for generative models to achieve; third, in generated riddles, the wording of some descriptions, while natural, does not contribute to riddle solving.

\section{Conclusion}
    We propose CC-Riddle, the first large scale dataset for Chinese Character Riddles covering most frequent simplified Chinese characters. Solving Chinese character riddles requires knowledge of character glyph, common sense (especially knowledge related to Chinese traditional culture), and understanding of rhetorical techniques. All these factors make the task challenging. Our experiments reveal that current language models fall short in these areas. Finally, we hope that CC-Riddle helps research on Chinese characters and figurative language.

\begin{ack}
We would like to convey our profound gratitude to all the annotators who have contributed to this research, with a special mention to Annotator A. Engaged in the manual riddle-solving test, this dedicated senior undergraduate majoring in Chinese linguistics shows remarkable commitment. We deeply appreciate their interest in our work. Through our interactions with the annotators, we are able to gain a richer understanding and a more refined perspective on the unique characteristics and challenges of Chinese character riddles.
\end{ack}

%\section*{References}
\bibliographystyle{abbrv}
\bibliography{CC-Riddle}

\begin{thebibliography}{10}

\bibitem{binsted1994implemented}
K.~Binsted and G.~Ritchie.
\newblock An implemented model of punning riddles.
\newblock Technical report, University of Edinburgh, Department of Artificial
  Intelligence, 1994.

\bibitem{binsted1994symbolic}
K.~Binsted and G.~Ritchie.
\newblock A symbolic description of punning riddles and its computer
  implementation.
\newblock {\em arXiv preprint cmp-lg/9406021}, 1994.

\bibitem{chiang2021transcouplet}
K.-Y. Chiang, S.~Lin, J.~Chen, Q.~Yin, and Q.~Jin.
\newblock Transcouplet: Transformer based chinese couplet generation.
\newblock {\em arXiv preprint arXiv:2112.01707}, 2021.

\bibitem{devlin2018bert}
J.~Devlin, M.-W. Chang, K.~Lee, and K.~Toutanova.
\newblock Bert: Pre-training of deep bidirectional transformers for language
  understanding.
\newblock {\em arXiv preprint arXiv:1810.04805}, 2018.

\bibitem{du2022glm}
Z.~Du, Y.~Qian, X.~Liu, M.~Ding, J.~Qiu, Z.~Yang, and J.~Tang.
\newblock Glm: General language model pretraining with autoregressive blank
  infilling.
\newblock In {\em Proceedings of the 60th Annual Meeting of the Association for
  Computational Linguistics (Volume 1: Long Papers)}, pages 320--335, 2022.

\bibitem{efrat2021cryptonite}
A.~Efrat, U.~Shaham, D.~Kilman, and O.~Levy.
\newblock Cryptonite: A cryptic crossword benchmark for extreme ambiguity in
  language.
\newblock In {\em Proceedings of the 2021 Conference on Empirical Methods in
  Natural Language Processing}, pages 4186--4192, 2021.

\bibitem{galvan2016riddle}
P.~Galv{\'a}n, V.~Francisco, R.~Herv{\'a}s, and G.~M{\'e}ndez.
\newblock Riddle generation using word associations.
\newblock In {\em Proceedings of the Tenth International Conference on Language
  Resources and Evaluation (LREC'16)}, pages 2407--2412, 2016.

\bibitem{lewis2019bart}
M.~Lewis, Y.~Liu, N.~Goyal, M.~Ghazvininejad, A.~Mohamed, O.~Levy, V.~Stoyanov,
  and L.~Zettlemoyer.
\newblock Bart: Denoising sequence-to-sequence pre-training for natural
  language generation, translation, and comprehension.
\newblock {\em arXiv preprint arXiv:1910.13461}, 2019.

\bibitem{li2015component}
Y.~Li, W.~Li, F.~Sun, and S.~Li.
\newblock Component-enhanced chinese character embeddings.
\newblock In {\em Proceedings of the 2015 Conference on Empirical Methods in
  Natural Language Processing}, pages 829--834, 2015.

\bibitem{li2021glyphcrm}
Y.~Li, Y.~Zhao, B.~Hu, Q.~Chen, Y.~Xiang, X.~Wang, Y.~Ding, and L.~Ma.
\newblock Glyphcrm: Bidirectional encoder representation for chinese character
  with its glyph.
\newblock {\em arXiv preprint arXiv:2107.00395}, 2021.

\bibitem{lin2021riddlesense}
B.~Y. Lin, Z.~Wu, Y.~Yang, D.-H. Lee, and X.~Ren.
\newblock Riddlesense: Reasoning about riddle questions featuring linguistic
  creativity and commonsense knowledge.
\newblock In {\em Findings of the Association for Computational Linguistics:
  ACL-IJCNLP 2021}, pages 1504--1515, 2021.

\bibitem{lu2016multi}
Y.~Lu, Y.~Zhang, and D.~Ji.
\newblock Multi-prototype chinese character embedding.
\newblock In {\em Proceedings of the tenth international conference on language
  resources and evaluation (LREC'16)}, pages 855--859, 2016.

\bibitem{papineni2002bleu}
K.~Papineni, S.~Roukos, T.~Ward, and W.-J. Zhu.
\newblock Bleu: a method for automatic evaluation of machine translation.
\newblock In {\em Proceedings of the 40th annual meeting of the Association for
  Computational Linguistics}, pages 311--318, 2002.

\bibitem{reimers2019sentence}
N.~Reimers and I.~Gurevych.
\newblock Sentence-bert: Sentence embeddings using siamese bert-networks.
\newblock In {\em Proceedings of the 2019 Conference on Empirical Methods in
  Natural Language Processing and the 9th International Joint Conference on
  Natural Language Processing (EMNLP-IJCNLP)}, pages 3982--3992, 2019.

\bibitem{shao2021cpt}
Y.~Shao, Z.~Geng, Y.~Liu, J.~Dai, F.~Yang, L.~Zhe, H.~Bao, and X.~Qiu.
\newblock Cpt: A pre-trained unbalanced transformer for both chinese language
  understanding and generation.
\newblock {\em arXiv preprint arXiv:2109.05729}, 2021.

\bibitem{sun2021chinesebert}
Z.~Sun, X.~Li, X.~Sun, Y.~Meng, X.~Ao, Q.~He, F.~Wu, and J.~Li.
\newblock Chinesebert: Chinese pretraining enhanced by glyph and pinyin
  information.
\newblock In {\em Proceedings of the 59th Annual Meeting of the Association for
  Computational Linguistics and the 11th International Joint Conference on
  Natural Language Processing (Volume 1: Long Papers)}, pages 2065--2075, 2021.

\bibitem{tan2016solving}
C.~Tan, F.~Wei, L.~Dong, W.~Lv, and M.~Zhou.
\newblock Solving and generating chinese character riddles.
\newblock In {\em Proceedings of the 2016 Conference on Empirical Methods in
  Natural Language Processing}, pages 846--855, 2016.

\bibitem{tseng2019eigencharacter}
Y.-H. Tseng and S.-K. Hsieh.
\newblock Eigencharacter: An embedding of chinese character orthography.
\newblock In {\em Proceedings of the Beyond Vision and LANguage: inTEgrating
  Real-world kNowledge (LANTERN)}, pages 24--28, 2019.

\bibitem{wolf2020transformers}
T.~Wolf, L.~Debut, V.~Sanh, J.~Chaumond, C.~Delangue, A.~Moi, P.~Cistac,
  T.~Rault, R.~Louf, M.~Funtowicz, et~al.
\newblock Transformers: State-of-the-art natural language processing.
\newblock In {\em Proceedings of the 2020 conference on empirical methods in
  natural language processing: system demonstrations}, pages 38--45, 2020.

\bibitem{xiang2008chinese}
C.~Xiang-li.
\newblock Chinese character rhetoric means as embodied in logogriph.
\newblock {\em Journal of Xiangnan University}, 2008.

\bibitem{yi2017generating}
X.~Yi, R.~Li, and M.~Sun.
\newblock Generating chinese classical poems with rnn encoder-decoder.
\newblock In {\em Chinese Computational Linguistics and Natural Language
  Processing Based on Naturally Annotated Big Data}, pages 211--223. Springer,
  2017.

\bibitem{zhang2021birdqa}
Y.~Zhang and X.~Wan.
\newblock Bird{QA}: A bilingual dataset for question answering on tricky
  riddles.
\newblock {\em arXiv preprint arXiv:2109.11087}, 2021.

\end{thebibliography}

\newpage

\appendix

\section{Appendix}
\subsection{Glyph Decomposition}
    \begin{figure}[H]
    \centering
    \includegraphics[width=\linewidth]{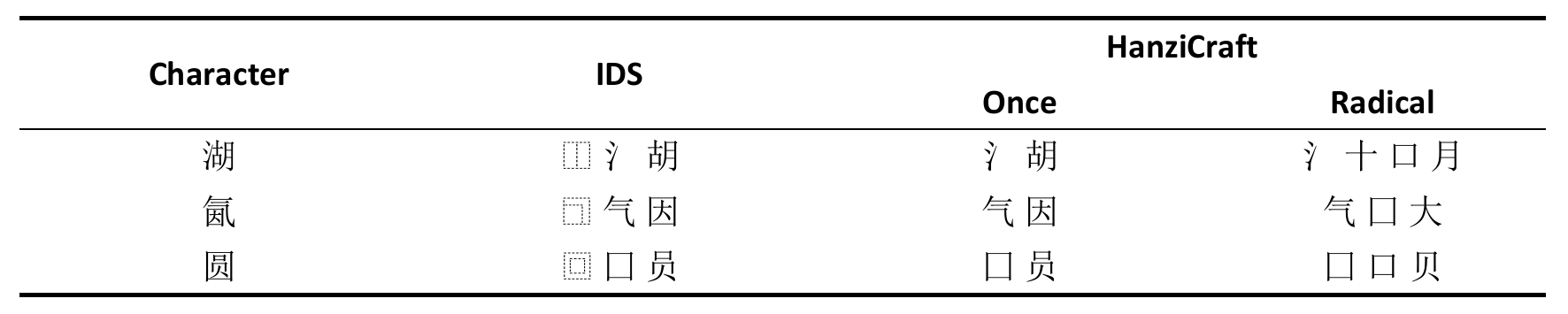}
    \caption{Examples of three different ways to decompose Chinese characters.} 
    \label{fig.decomposition}
    \end{figure}

\subsection{Manual Annotation}
    \label{appendix:annotation}
    We offer a fair compensation to our annotators. For manual filtering, each annotator is compensated with 0.1 yuan for every candidate they review. For manual test on solving the 100 sampled riddles, we offer a reward of 200 yuan to each annotator. We recognize the extra effort of Annotator A by paying an additional 60 yuan.

\subsection{Computational Cost}
    \label{appendix:computational_cost}
    The riddle generation experiment is conducted on a single 1080Ti GPU, which performs approximately 4e+16 FLOPS. The retrieval-based QA experiments is finished on an A40 GPU within an hour. ChatGLM is also operated on a single A40 GPU, and all experiments using ChatGLM take a total of roughly 3 hours. For ChatGPT, we utilize the OpenAI API and use a total of 665,561 tokens.

\end{CJK}
\end{document}